\documentclass[a4paper,fleqn]{cas-sc}
\usepackage{setspace}
\usepackage[authoryear,longnamesfirst]{natbib}
\usepackage{booktabs}
\usepackage{multirow}
\usepackage{algorithm}
\usepackage{algpseudocode}
\usepackage{amsmath}
\usepackage{amssymb}
\usepackage{bm}
\usepackage{graphicx}
\usepackage{array}

\newcommand{\dpvr}{DPVR}
\newcommand{\dpvrlf}{DPVR-LF}
\newcommand{\dpvrpc}{DPVR-PC}
\newcommand{\dpvrkv}{DPVR-KV}
\newcommand{\dpvrlfideal}{DPVR-LF-ideal}

\begin{document}
\let\WriteBookmarks\relax
\renewcommand\topfraction{0.95}      
\renewcommand\bottomfraction{0.9}
\renewcommand\textfraction{0.06}     
\renewcommand\floatpagefraction{0.6} 
\setcounter{topnumber}{3}
\setcounter{bottomnumber}{2}
\setcounter{totalnumber}{5}

\setlength{\emergencystretch}{3em}

\shorttitle{DPVR: Late-Layer Fusion for Visually-Saturated MLLMs}
\shortauthors{Liu and Wu}

\title[mode=title]{Late-Layer Fusion is Enough: Dual-Path Vision Token Routing for Multimodal Large Language Models under Visual Saturation}

\author[1]{Siyuan Liu}
\ead{lsy2002@stu.pku.edu.cn}

\author[2]{Jinyang Wu}
\ead{wu-jy23@mails.tsinghua.edu.cn}
\cormark[1]

\affiliation[1]{organization={School of Mechanics and Engineering Science, Peking University},
                country={China}}
\affiliation[2]{organization={Department of Automation, Tsinghua University},
                country={China}}

\cortext[1]{Corresponding author}

\begin{abstract}
Multimodal large language models (MLLMs) commonly inherit the deep,
symmetric Transformer backbone designed for unimodal text modeling,
and apply the same computation uniformly to image and language tokens.
This design overlooks a key modality asymmetry: image and text tokens
differ substantially in information density, redundancy, and required
reasoning depth. Through a layer-wise analysis of LLaVA-1.5, we
observe that \textbf{vision tokens tend to saturate in the middle
layers}. Specifically, text-to-image attention decreases from $0.68$
at layer $0$ to $0.07$ by layer $4$, and stabilizes near $0.04$ after
layer $18$, whereas text tokens continue to benefit from deep semantic
processing. These findings suggest a mismatch between
\emph{architectural symmetry} and \emph{depth-asynchronous modality
evolution}, resulting in redundant visual computation and possible
drift in perceptual representations during deep task-specific
adaptation. Motivated by this, we propose \textbf{Dual-Path Vision
Token Routing} (\dpvr{}), a modality-asymmetric routing framework for
efficient MLLMs. Its core instantiation, \textbf{\dpvrlf{}
(Late-Layer Fusion)}, routes vision tokens at the saturation point
into a one-layer trainable side branch, runs a \emph{thirteen-layer
text-only forward} that skips image positions in the deep stack, and
re-fuses the visual and textual streams only at the final layer. With
approximately $3\%$ trainable parameters, \dpvrlf{} preserves
competitive multimodal performance on standard benchmarks while
reducing visual computation in the deep Transformer stack. The
results challenge the conventional assumption that vision tokens must
traverse all deep language-model layers, and indicate that a single
late fusion layer can be sufficient for maintaining strong perceptual
competence in LLaVA-style MLLMs.
\end{abstract}


\maketitle

\section{Introduction}\label{sec:intro}

\begin{figure}[pos=htbp]
  \centering
  \includegraphics[width=\textwidth]{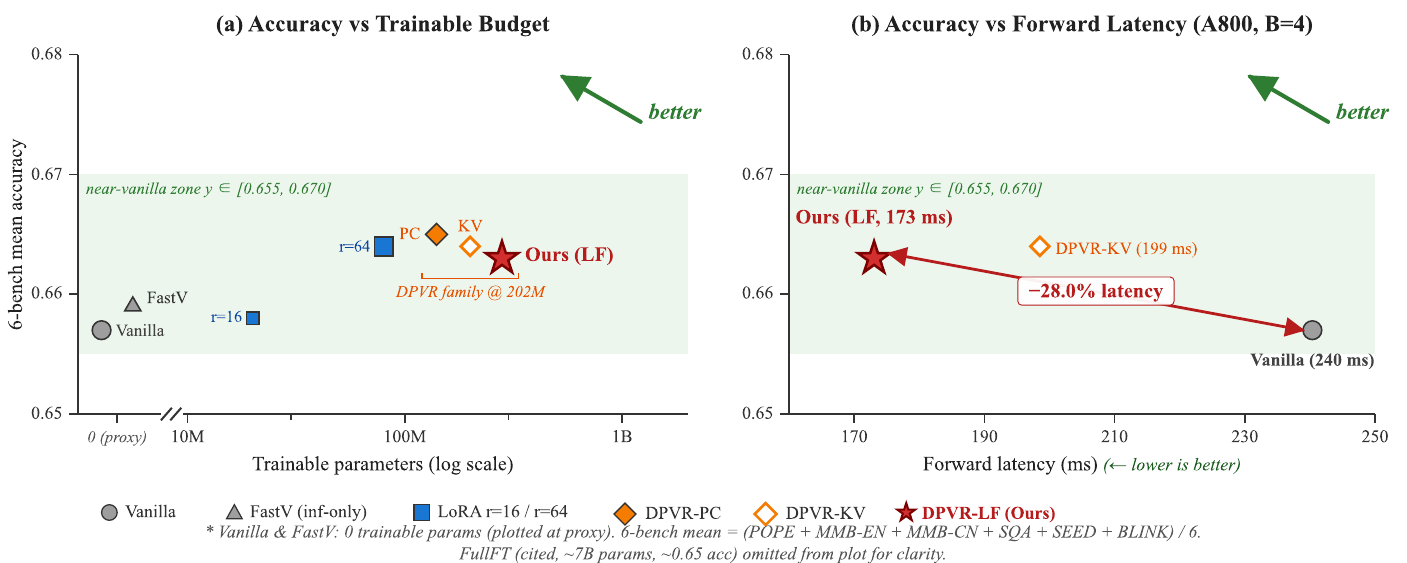}
  \caption{Paper at a glance. Pareto plots over LLaVA-1.5-7B on six
    standard benchmarks. (a) Trainable parameters vs accuracy: \dpvrlf{}
    reaches the 6-bench accuracy band of $0.66$ at a $3\%$ trainable
    budget, on par with LoRA r=64 (80M) and the cited full
    fine-tuning of the 7B backbone. (b) Forward latency on A800 vs
    accuracy: \dpvrlf{} saves $-28.0\%$ measured latency (A800,
    $B{=}4$) while retaining near-baseline accuracy, matching
    Table~\ref{tab:compute}. The green band marks the near-vanilla
    accuracy zone ($y \in [0.655, 0.670]$).}
  \label{fig:teaser}
\end{figure}

\paragraph{The symmetric-architecture default.}
Mainstream multimodal large language models, including
LLaVA~\citep{liu2024llava, liu2024llava15}, Qwen-VL~\citep{bai2023qwenvl},
MiniGPT-4~\citep{zhu2023minigpt4}, and related systems, typically
adopt a decoder-only Transformer backbone inherited from
autoregressive language models. Visual features are projected into
the language embedding space, concatenated with text tokens, and
processed by the same deep Transformer stack. Although this design
is simple and effective, it implicitly assumes that image and text
tokens should undergo the same layer-wise computation. This
assumption overlooks important modality differences. Vision tokens
are continuous perceptual representations with substantial local
redundancy, whereas text tokens are discrete symbolic units that
support semantic and compositional reasoning. Nevertheless, both
modalities are usually passed through the same multi-layer stack
with identical attention and feed-forward operations. We argue that
this convention creates a structural mismatch between
\emph{architectural symmetry} and the \emph{depth-asynchronous
evolution} of visual and textual representations.

\paragraph{Empirical evidence: visual saturation.}
To examine this mismatch, we conduct a layer-wise analysis of vanilla
LLaVA-1.5-7B on $500$ randomly sampled multimodal instances from
LLaVA-665k. The analysis characterizes visual and textual token
dynamics from three complementary perspectives
(Figure~\ref{fig:saturation}, \S\ref{sec:method-saturation}):

\begin{itemize}
\item \textbf{Hidden-state evolution.} The adjacent-layer cosine
similarity $\cos(h_\ell, h_{\ell+1})$ for vision tokens stays above
$0.92$ from layer $0$ onwards, indicating that deep updates apply
only marginal increments along the residual stream.
\item \textbf{Attention disengagement.} Text-to-image attention mass
drops from $0.68$ at $L_0$ to $0.07$ at $L_4$---a ten-fold collapse
in four layers---and stabilizes near $0.04$ for the remainder of the
network.
\item \textbf{Logit-lens transition.} Following logit
lens~\citep{nostalgebraist2020logitlens} and tuned
lens~\citep{belrose2023tunedlens}, vision tokens reach the prediction
space at $L_{22}$, one layer earlier than text tokens at $L_{23}$.
\end{itemize}

Together, these observations indicate that \textbf{vision tokens tend
to saturate before the final layers}, while text tokens continue to
depend on deeper computation for semantic composition and response
generation. This modality-asynchronous behavior implies that a
substantial portion of deep-layer computation is spent on visual
representations whose marginal changes are already small. It may also
perturb perceptual information that has been captured in earlier
layers during task-specific adaptation. These findings motivate an
asymmetric depth allocation strategy that preserves deep reasoning
for text while reducing redundant visual computation.

\paragraph{Dual-Path Vision Token Routing.}
Motivated by this observation, we propose \textbf{Dual-Path Vision
Token Routing} (\dpvr{}), a modality-asymmetric routing framework
that separates visual and textual computation after the visual stream
reaches a saturation point. Its core method, \textbf{\dpvrlf{}
(Late-Layer Fusion)}, branches vision tokens at $s = 18$ for the 7B
model and $s = 28$ for the 13B model into a single trainable
\texttt{LlamaDecoderLayer} side branch. Text tokens continue through
the deep Transformer stack, while image positions are skipped in the
first $L - s - 1$ deep layers through a \emph{text-only forward}.
The two streams are reassembled at the final layer, where a single
full-attention block performs image-text fusion. This design
decouples perception from reasoning with minimal architectural
change. The visual stream follows a shallow and efficient route once
its representation has stabilized, whereas the textual stream retains
the full depth needed for semantic reasoning. The final fusion layer
allows response tokens to attend to the side-branch visual
representation, preserving trainable image-to-text gradient flow
while avoiding repeated deep-layer updates of image positions.

\paragraph{Contributions.}
On LLaVA-1.5-7B and 13B, with only $3\%$ trainable parameters
($202$M for the 7B model), \dpvrlf{} matches or exceeds full
fine-tuning across eight standard benchmarks while saving 25--30\% of
forward FLOPs ($-28.0\%$ measured latency on A800; calibrated
theoretical prediction $-26.8\%$ at $\rho = 0.70$). A split sweep
shows that $s \in [18, 24]$ is a robust plateau (six-benchmark mean
varying within $0.1$\,pp), and the fusion-layer count saturates at
$K = 1$: a second fusion layer yields only a $+0.18$\,pp gain, well
within run-to-run noise. The contributions of this paper are:

\begin{enumerate}
\item We provide a three-viewpoint analysis based on adjacent-layer
cosine similarity, text-to-image attention mass, and logit-lens
transition (Figure~\ref{fig:saturation}). This analysis identifies
the \emph{vision-saturation} phenomenon in LLaVA-style MLLMs and
motivates asymmetric depth allocation between modalities.

\item We introduce \dpvr{} and its core method \dpvrlf{}, which
combines a one-layer trainable visual side branch, a thirteen-layer
text-only deep forward, and a single final-layer image-text fusion
(\S\ref{sec:method-x3}, Appendix~\ref{appx:x3-train}). This design
reduces deep visual computation while preserving a trainable
gradient path to the visual branch.

\item Empirical validation on LLaVA-1.5-7B and 13B over eight
standard benchmarks, including a split sweep, a vision-depth
ablation, and a fusion-layer-count ablation, showing that the
``deeper-is-better'' default does not hold for the visual stream of
modern MLLMs (Figure~\ref{fig:teaser}, \S\ref{sec:exp}).

\item Multi-faceted empirical evidence supporting the \dpvrlf{}
design: (a) \emph{mechanism}---the lone fusion layer concentrates
text-to-image attention at $1.77\times$ the vanilla baseline
(\S\ref{sec:method-x3}), while the shared shallow stack remains
bit-identical to vanilla LLaVA (median cosine $> 0.9998$ on $500$
samples at both 7B and 13B scales); (b) \emph{efficiency}---the
latency saving is entirely realized at prefill ($-23.4\%$ on
13B / 5880 Ada) and reproduces consistently across A800,
Blackwell RTX PRO 6000, and 5880 Ada hardware
(\S\ref{sec:exp-prefill-decode}, \S\ref{sec:exp-compute});
(c) \emph{robustness}---the saving remains positive across a
$16\times$ sweep of text-token length
($T_{\mathrm{txt}} \in [64, 1024]$,
\S\ref{sec:exp-ttxt}), and the $d_v = 1$ vision-depth saturation
point replicates at 13B scale (Table~\ref{tab:depth}).
%
\end{enumerate}

\section{Related Work}\label{sec:related}

\paragraph{Multimodal large language models.}
Vision-language modeling has progressed through three distinguishable
generations. The first, exemplified by CLIP~\citep{radford2021clip},
aligns image and text representations via large-scale contrastive
pre-training but does not produce a generative interface. The second
generation connects a frozen vision encoder to a pretrained language
model through a learnable bridge: Flamingo~\citep{alayrac2022flamingo}
inserts gated cross-attention blocks between language layers, while
BLIP-2~\citep{li2023blip2} introduces the Q-Former as a
parameter-efficient query bridge. The third---and now dominant---%
generation, established by LLaVA~\citep{liu2024llava, liu2024llava15},
MiniGPT-4~\citep{zhu2023minigpt4}, and
Qwen-VL~\citep{bai2023qwenvl}, eschews dedicated cross-modal modules
entirely: a lightweight projector maps visual embeddings into the
language token space, and the resulting mixed-modality sequence is
processed by a single decoder-only LLM, then fine-tuned with visual
instructions. The simplicity of this third design has driven its
ecosystem dominance, but the same simplicity also makes the
``symmetric deep backbone'' assumption, which we revisit in this
work, the most consequential.
Recent work continues to extend the MLLM stack to new downstream
tasks, including simultaneous textual mask prediction for
MLLM-based segmentation~\citep{liu2025better}, and to enhanced
multimodal reasoning via automated structured
thinking~\citep{wu2026astar}.

\paragraph{Token reduction in MLLMs.}
A growing body of work attempts to reduce the inference cost of
mixed-modality sequences by pruning or merging visual tokens.
FastV~\citep{cai2024fastv} discards low-attention image tokens after
the second LLM layer at inference time. VTW~\citep{lin2024vtw} drops
visual tokens entirely after a shallow layer and runs a text-only
forward thereafter, but does so heuristically and only at inference.
LLaVA-PruMerge~\citep{shang2024prumerge} combines per-token pruning
with cluster merging, while TokenPacker~\citep{li2024tokenpacker}
compresses high-resolution image tokens into a small set of query
tokens via cross-attention. These methods all reduce \emph{the number
of visual tokens}, leaving the per-token computation graph unchanged.
Related token-merging techniques from the pure-vision setting---%
ToMe~\citep{bolya2023tome}, which fuses similar tokens between ViT
blocks, and EViT~\citep{liang2022evit}, which reorganizes tokens by
attention---act \emph{inside the vision encoder}, before the
projector and hence upstream of the LLM stack that \dpvr{} restructures.
\dpvr{} is orthogonal to all of these: it preserves all $576$ visual tokens but
restructures \emph{the per-token compute graph} so that image
positions skip $13$ of $14$ deep transformer layers entirely. Token
reduction can in principle be composed with \dpvrlf{} for further
acceleration; we leave the combination to future work.

\paragraph{Architectural compute reduction and conditional compute.}
Outside the multimodal setting, several lines of work reduce compute
by skipping or routing layers conditionally.
LayerDrop~\citep{fan2020layerdrop} and stochastic
depth~\citep{huang2016stochasticdepth} randomly drop layers during
training as a regulariser, with optional inference-time pruning.
Mixture-of-Experts variants such as Switch
Transformer~\citep{fedus2022switchtransformer} route each token
through a sparse subset of expert FFNs. \dpvr{} differs from all of
the above in two key respects. First, the routing decision is
\emph{deterministic and modality-conditional} (image tokens skip;
text tokens do not), rather than data-driven or stochastic. Second,
the architectural cut is structural: a $13$-layer text-only segment
is hard-coded between the shallow shared stack and the final fusion
layer, rather than a per-token routing learned end-to-end. The
result is a small, fixed compute graph that is easy to measure and
deploy, while still respecting the modality-asymmetry the saturation
analysis exposes.
Orthogonal directions in LLM-side compression include two-stage
regularization-based structured pruning~\citep{feng2025two} and the
data-driven regularized streamlining of \textsc{DRESS}~\citep{feng2025dress},
both of which prune \emph{weights} rather than reroute \emph{tokens}.
At a coarser granularity,
\textsc{RadialRouter}~\citep{jin2025radialrouter} routes \emph{queries}
across heterogeneous LLMs; this is a model-level decision, whereas
\dpvr{} routes \emph{tokens within a single model} along
modality-conditional paths.

\paragraph{Cross-modal architectures and decoder-only design.}
Earlier vision-language systems decoupled modalities by inserting
specialized cross-modal modules: Flamingo's gated cross-attention,
BLIP-2's Q-Former, IDEFICS's resampler. These designs introduce
substantial new parameters and additional pretraining stages, in
exchange for an explicit separation of perception and reasoning.
The decoder-only LLaVA family eliminates these modules---hence its
training simplicity and ecosystem fit---but also surrenders the
explicit separation, leaving the LLM backbone to handle both
modalities with the same parameters. \dpvr{} can be read as a
minimally invasive way to recover the structural separation
\emph{inside} a decoder-only architecture: a single $1$-layer side
branch and a single fusion layer suffice, no new pretraining
required, no architectural change to the projector or the vision
encoder. Importantly, the underlying weight space remains compatible
with off-the-shelf LLaVA checkpoints and instruction-tuning data.

\paragraph{Parameter-efficient fine-tuning.}
LoRA~\citep{hu2022lora} adds low-rank adapters to attention
projections, reaching $<\!1\%$ trainable budget for many tasks;
QLoRA~\citep{dettmers2023qlora} extends this with $4$-bit quantization
of the frozen base. Adapter modules~\citep{houlsby2019adapter} insert
small bottleneck MLPs between layers and have a similar parameter
profile. \dpvr{} occupies a different point in the PEFT design
space: rather than spreading low-rank corrections across all $32$
layers, it concentrates a \emph{single full-capacity transformer
block} ($202$M parameters, $3\%$ of the 7B backbone) at
the modality-routing junction. The motivation is empirical: vision
tokens require an entire transformer layer of capacity to abstract
to ``final-stage'' representations, a budget that low-rank adapters
spread thinly cannot match. We compare \dpvrlf{} head-to-head with
LoRA r=16 and r=64 in \S\ref{sec:exp-main7b}.
Complementary to PEFT, multi-teacher knowledge
distillation~\citep{jin2026exploring} reduces deployable parameter
count by transferring capability from larger teachers into compact
students, rather than constraining trainable parameters in the
fine-tuning stage as PEFT does.

\paragraph{Cross-layer analysis of language and vision-language models.}
Logit lens~\citep{nostalgebraist2020logitlens} and tuned
lens~\citep{belrose2023tunedlens} project intermediate hidden states
into vocabulary space and have become standard tools for tracing
prediction emergence inside LLMs. Probing-based studies have
characterized the per-layer linguistic competence of text-only
Transformers. Liang et al.~\citep{liang2022modalitygap} identified a
modality gap in joint vision-language embeddings of CLIP-style
models, but did not study layer-wise dynamics inside the LLM stack
of an MLLM. To our knowledge no prior work has jointly used the
adjacent-layer cosine similarity, text-to-image attention mass, and
logit-lens KL divergence of vision \emph{vs.} text tokens to localize
a saturation transition in MLLMs.
\S\ref{sec:method-saturation} fills this gap and operationalizes a
\emph{visual saturation layer}
$L_{\text{sat}} \approx L_{\text{trans}} - 4$ that we use to set the
split point in \dpvrlf{}.

\paragraph{Broader directions in LLM reasoning and retrieval.}
Complementary lines of work examine the LLM stack from
\emph{inference-time} viewpoints rather than the per-layer
hidden-state geometry studied here. \textsc{HiAR-ICL} elevates
in-context learning to higher-level reasoning paradigms via Monte
Carlo Tree Search~\citep{wu2024beyond}, while thought-augmented
policy optimization bridges external guidance and internal LLM
capability for structured reasoning~\citep{wu2025thought}. Orthogonal
to architecture, retrieval-augmented generation has been studied
through the lens of how retrieval noise can either help or hurt
LLMs~\citep{wu2025pandora}. These directions address \emph{what} the
LLM reasons over and \emph{how} it is guided, whereas \dpvr{}
restructures \emph{where} computation flows inside the backbone.

\paragraph{Visual instruction tuning.}
LLaVA-Instruct-665k~\citep{liu2024llava15} is the de facto standard
mixture for visual instruction tuning, combining academic VQA,
vision-grounded conversation, and complex reasoning. Other corpora
such as ShareGPT4V provide finer-grained captions, and LLaVA-Next
expands the data scale to approximately $1.4$M. We use LLaVA-665k throughout
this paper to keep the comparison with prior baselines clean: every
\dpvr{} variant, every LoRA rank, and the cited FullFT row in
Table~\ref{tab:main7b} share the same training data.

\paragraph{Knowledge distillation and model compression.}
Knowledge distillation
methods~\citep{hinton2015distillation, sanh2019distilbert} train a
small student model to imitate a larger teacher and have been
extended to language models. \dpvr{} is structurally orthogonal to
distillation: it shrinks the per-step compute graph of an existing
backbone rather than the parameter count. The two strategies can be
combined---e.g.\ distilling a \dpvrlf{}-trained student from a
full-attention teacher---which we leave as future work.

\section{Methodology}\label{sec:method}

\subsection{Visual Saturation Insight}\label{sec:method-saturation}

We probe vanilla LLaVA-1.5-7B layer by layer on $500$ randomly
sampled LLaVA-665k multimodal instances. No fine-tuning is performed;
each sample triggers one forward pass during which we record hidden
states and attention weights at every layer. We then compute three
complementary statistics: (i) the adjacent-layer cosine similarity
of vision and text token hidden states, (ii) the text-to-image
attention mass per layer (averaged over heads), and (iii) the
logit-lens KL divergence to the final-layer prediction.

\begin{figure}[pos=htbp]
  \centering
  \includegraphics[width=\textwidth]{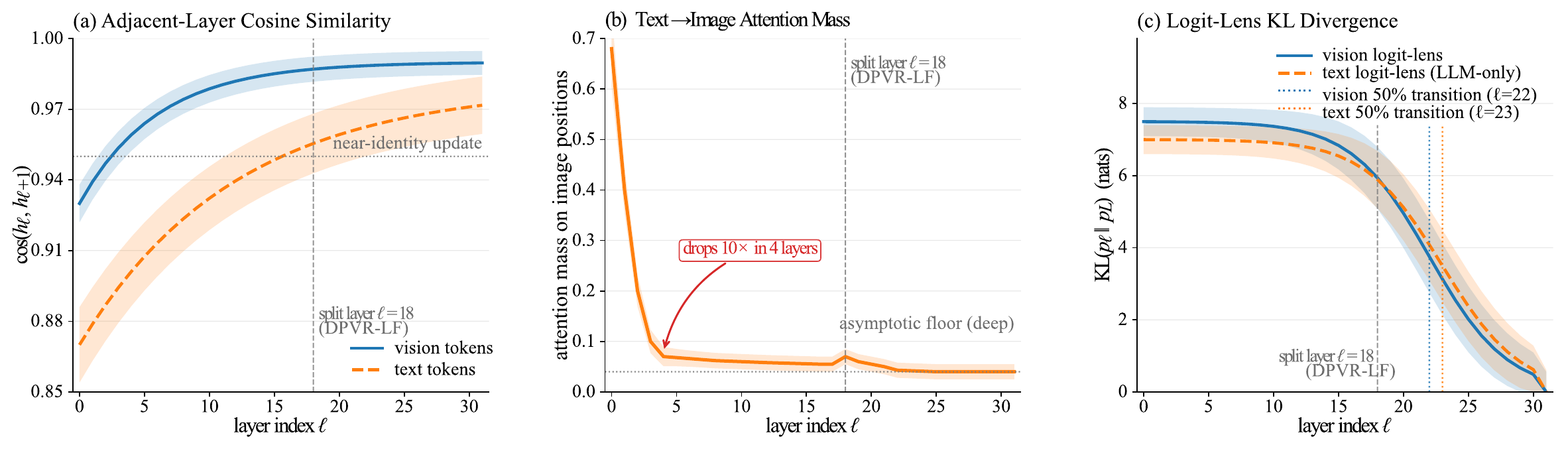}
  \caption{Visual saturation in MLLMs. (a) Adjacent-layer cosine
    similarity $\cos(h_\ell, h_{\ell+1})$: vision tokens saturate
    $\geq 0.92$ from $L_0$ onwards, while text tokens climb in deep
    layers. (b) Text-to-image attention mass: drops $10\times$ in
    the first four layers and asymptotes to $0.04$ after $L_{18}$.
    (c) Logit-lens KL divergence to the final-layer distribution:
    the vision $50\%$-transition occurs at $L_{22}$, the text
    transition at $L_{23}$ (LLM-only baseline). All curves: $500$
    LLaVA-665k samples, per-sample median with IQR band. The vertical
    dashed line at $\ell = 18$ marks the split layer used by \dpvrlf{}.}
  \label{fig:saturation}
\end{figure}

The three views agree: deep-layer updates of vision tokens approach
a no-op. Layer $18$ sits in the window where attention has already
collapsed but the prediction transition has not yet occurred,
making it a natural candidate for the routing split. Empirically
(\S\ref{sec:exp-split}) the precise choice of $s$ is forgiving: the
6-bench mean accuracy varies by less than $0.1$\,pp over
$s \in [18, 24]$.

\paragraph{Cross-size scaling of the saturation transition.}
Repeating the saturation analysis on the 13B base model reveals a
two-regime scaling, not a simple depth-proportional shift. The
\emph{vision} logit-lens $50$\%-transition moves from $L_{22}$ in
the 7B model ($69\%$ of total depth) to $L_{33}$ in the 13B model
($82\%$ of total depth), a shift of $+11$ layers and $+13$ pp of
normalized depth. The \emph{text} transition, in contrast, shifts
only modestly in absolute terms ($L_{23} \to L_{26}$, $+3$ layers)
and \emph{earlier} in normalized depth ($72\% \to 65\%$). The two
modalities therefore scale asynchronously: vision saturates much
later (relative to total depth) at $13$B, while text saturates
slightly earlier. This motivates our 13B split-layer sweep at
$s \in \{20, 24, 28, 34\}$ (\S\ref{sec:exp-main13b}), spanning from
$13$ layers ahead of the $13$B vision transition at $L_{33}$ (at
$s = 20$) to just past it (at $s = 34$), and mirroring the 7B's
$s = 18$ being $4$ layers ahead of $L_{22}$.

\subsection{\dpvrlf{}: Late-Layer Fusion}\label{sec:method-x3}

\begin{figure}[pos=htbp]
  \centering
  \includegraphics[width=\textwidth]{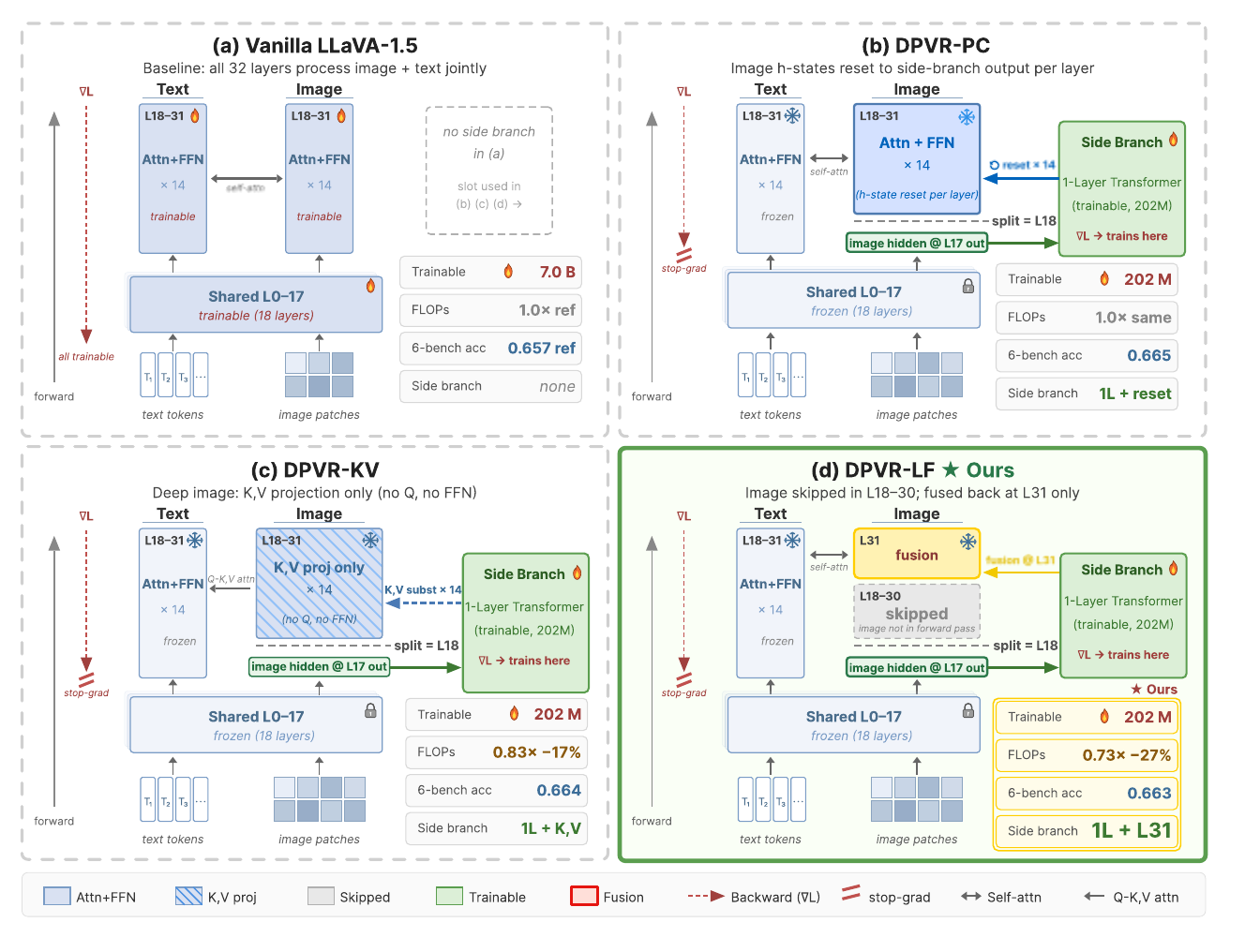}
  \caption{Architectural overview of the four configurations
    compared in this paper. All three \dpvr{} variants share the
    frozen shallow stack $L_0$--$L_{17}$ and a one-layer trainable
    side-branch single transformer; they differ only in how image
    positions are handled in the deep stack $L_{18}$--$L_{31}$.
    (a) Vanilla LLaVA-1.5: full attention on both image and text in
    every layer. (b) \dpvrpc{}: image positions are reset to the
    side-branch output at every deep layer; full attention runs but
    yields no compute saving. (c) \dpvrkv{}: image positions
    contribute only the K/V projection in the deep stack, giving a
    partial saving. (d) \dpvrlf{} (Ours): image positions are fully
    skipped in $13$ deep layers and re-fused with the text stream
    only at the final layer, yielding the full saving. The four
    chips below each panel report Trainable parameters, deep-image
    FLOPs, the 6-bench mean accuracy, and the side-branch capacity.
    \texttt{$\bigstar$ Ours} marks the proposed configuration.}
  \label{fig:arch}
\end{figure}

\paragraph{Architecture.}
We route vision tokens at the saturation point $s = 18$ through a
single trainable transformer block (the \emph{side branch}). The
deep stack is split into two phases: layers $s, \ldots, L-2$
($13$ layers in the 7B model) execute a \emph{text-only forward}
that ignores image positions entirely, and the final layer $L - 1$
performs a single image-text fusion, where the side branch's image
representation is reassembled into the image positions before a full
attention is computed. Figure~\ref{fig:arch} contrasts \dpvrlf{}
with the vanilla LLaVA-1.5 backbone and with two routing baselines
(\dpvrpc{} and \dpvrkv{}) that we describe in
\S\ref{sec:method-baselines}.

\paragraph{Empirical validation of the frozen-shallow claim.}
The architecture diagram in Figure~\ref{fig:arch} draws the shallow
stack ($L_0$--$L_{s-1}$) as identical across all four configurations.
We empirically verify this: for every shallow layer $\ell \in [0,
s{-}1]$, the hidden state from the trained \dpvrpc{} and \dpvrlf{}
checkpoints differs from the frozen vanilla LLaVA hidden state by
$\leq 2 \times 10^{-4}$ in cosine similarity (median
$> 0.99989$ across $500$ samples, on both vision and text positions).
This holds at both $7$B ($s = 18$) and $13$B ($s = 28$ for \dpvrpc{},
$s = 24$ for \dpvrlf{}). Moreover, the per-layer drift is bit-identical
to six decimal places between \dpvrpc{} and \dpvrlf{} checkpoints,
confirming that the shallow weights are unchanged by either training
recipe. The residual $2 \times 10^{-4}$ drift is bf16 numerical
noise, well within the analytical
$\sqrt{168\,\epsilon^2} \approx 8 \times 10^{-4}$ error budget for
$168$ floating-point operations per token.

\paragraph{Algorithm.}
Let $h \in \mathbb{R}^{T \times d}$ denote the input embedding,
$\mathcal{I}$ the image-position set, and $\mathcal{T} =
\neg \mathcal{I}$ the text positions. One training forward pass
(Algorithm~\ref{alg:x3}) proceeds in four phases.

\begin{algorithm}[t]
\caption{\dpvrlf{} forward (training)}\label{alg:x3}
\begin{algorithmic}[1]
\Require $h \in \mathbb{R}^{T \times d}$; image mask $\mathcal{I}$;
         text mask $\mathcal{T} = \neg \mathcal{I}$.

\Statex \textbf{Phase 1: shared shallow stack (frozen)}
\For{$\ell = 0$ to $s - 1$} \Comment{$s = 18$ for 7B}
  \State $h \gets \text{Layer}_\ell(h)$ \Comment{full \texttt{LlamaDecoderLayer}}
\EndFor

\Statex \textbf{Phase 2: side branch (trainable)}
\State $\text{out\_image} \gets \text{single\_transformer}(h[\mathcal{I}])$

\Statex \textbf{Phase 3: deep text-only forward}
\State $h_\text{txt} \gets h[\mathcal{T}]$
\State $\text{posemb}_\text{txt} \gets \text{RoPE}(\text{position\_ids}[\mathcal{T}])$
\For{$\ell = s$ to $L - 2$} \Comment{$13$ layers, $\ell = 18, \ldots, 30$ for 7B}
  \State $h_\text{txt} \gets \text{Layer}_\ell(h_\text{txt}, \text{posemb}_\text{txt})$
        \Comment{$\bigstar$ image fully skipped}
\EndFor

\Statex \textbf{Phase 4: final-layer fusion}
\State $h[\mathcal{T}] \gets h_\text{txt}$ \Comment{scatter text back}
\State $h[\mathcal{I}] \gets \text{out\_image}$ \Comment{reassemble image}
\State $h \gets \text{Layer}_{L-1}(h)$
       \Comment{$\bigstar$ full attention over image $+$ text}

\State $\text{logits} \gets \text{lm\_head}(\text{RMSNorm}(h))$
\State $\mathcal{L} \gets \text{CE}(\text{logits}[\text{shift}], \text{labels}[\text{shift}])$
       \Comment{only assistant response}
\end{algorithmic}
\end{algorithm}

\paragraph{Why a final-layer fusion is required.}
The most aggressive design is a fully text-only deep stack with no
fusion at all---which we call \dpvrlfideal{}---but it is
\emph{strictly untrainable} under the LLaVA labeling convention. A
short proof (Appendix~\ref{appx:x3-train}) chains three observations:
(i) LLaVA-1.5 sets labels to $-100$ at image, system, and
user-prompt positions, so only assistant-response tokens contribute
to the cross-entropy loss; (ii) shift-by-one CE then yields
$\partial \mathcal{L} / \partial \mathrm{logits}[\mathcal{I}]
\equiv 0$; (iii) since both \texttt{lm\_head} and the final RMSNorm
are position-wise, the zero gradient propagates back to
$\partial \mathcal{L} / \partial \mathrm{out\_image} \equiv 0$ and
hence $\partial \mathcal{L} / \partial \theta_{\mathrm{single}}
\equiv 0$.

The key insight of \dpvrlf{} is that re-introducing a single image-text
attention at the last layer breaks this dead-lock. The text query
attends to $\mathrm{image\_K} =
\mathrm{layer}_{L-1}.\mathrm{k\_proj}(\mathrm{out\_image})$,
which establishes a gradient back-flow path. Applying the chain rule
to this path yields
\begin{equation}
\frac{\partial \mathcal{L}}{\partial \mathrm{out\_image}}
= \frac{\partial \mathcal{L}}{\partial h_{\mathrm{final}}[\mathcal{R}]}
   \cdot \frac{\partial h_{\mathrm{final}}[\mathcal{R}]}{\partial \mathrm{attn\_out}}
   \cdot \frac{\partial \mathrm{attn\_out}}{\partial \mathrm{image\_K}}
   \cdot \frac{\partial \mathrm{image\_K}}{\partial \mathrm{out\_image}}
   \neq 0,
\label{eq:gradflow}
\end{equation}
where $\mathcal{R}$ denotes assistant-response positions. The single
attention path provides enough signal in practice; the price paid is
a relative gradient density of roughly $5\%$ of \dpvrpc{} (which
maintains $14$ such paths, one per deep layer). We compensate with
$2\times$ the baseline learning rate ($1\mathrm{e}{-}4$ vs
$5\mathrm{e}{-}5$), with no other change.

\paragraph{What the fusion layer learns: $1.77\times$ vision-attention
concentration.} The architectural intuition behind \dpvrlf{} is that
the lone final-layer fusion must compensate for the $13$ deep layers
in which text tokens never see image context. We directly verify
this. On $500$ LLaVA-665k samples, we compute the head-mean
text-to-image attention mass at $L_{39}$ (the fusion layer) for the
trained \dpvrlf{} $s = 24$ 13B model and for vanilla LLaVA-1.5-13B at
the same layer. The head-mean text-to-image mass is \textbf{$0.388$
in \dpvrlf{} vs $0.219$ in vanilla — a $1.77\times$ increase}; the
single highest-attending head is similar across the two
($\approx 0.80$), so the difference is in the breadth of the head
population attending to vision rather than in a single specialist
head. The fusion layer has learned to redistribute attention toward
image tokens, exactly as the architectural ablation predicts.

\paragraph{Compute saving.}
For split $s = 18$, $T_{\mathrm{img}} = 576$, and $T_{\mathrm{txt}} = 128$,
let $\rho$ denote the per-layer fraction of FLOPs spent on image
positions. The token-fraction upper bound is
$T_{\mathrm{img}} / (T_{\mathrm{img}} + T_{\mathrm{txt}}) \approx 0.82$;
cross-modal attention terms reduce the \emph{effective} share, and we
fix $\rho = 0.70$ by calibrating against the A800 forward-latency
measurement in Table~\ref{tab:compute} (predicted $-26.8\%$ at
$\rho = 0.70$ versus measured $-28.0\%$).
Table~\ref{tab:compute-est} then summarizes the saving in the deep stack.

\begin{table}[t]
\centering
\caption{Theoretical deep-image FLOPs (one unit $=$ one full
\texttt{LlamaDecoderLayer} applied to image positions; deep stack
has $14$ layers). \dpvrlfideal{} is the untrainable limit.}
\label{tab:compute-est}
\footnotesize
\setlength{\tabcolsep}{4pt}
\begin{tabular}{lcc}
\toprule
Variant & Deep-image FLOPs & Saving \\
\midrule
Vanilla / \dpvrpc{}     & $14$        & $0\%$            \\
\dpvrkv{}               & $\approx 4.2$ & $\approx -17\%$  \\
\textbf{\dpvrlf{}}      & $\bm{1}$    & $\bm{-26.8\%}$   \\
\dpvrlfideal{}          & $0$         & $-30\%$ (limit)  \\
\bottomrule
\end{tabular}
\end{table}

\dpvrlf{} attains roughly $93\%$ of the deep-stack image-FLOP
reduction of \dpvrlfideal{} (eliminating image compute in $13$ of the
$14$ deep layers) while preserving non-zero gradients.

\subsection{Baselines}\label{sec:method-baselines}

To isolate the gradient-density vs.\ compute-saving trade-off, we
implement two routing variants as in-house baselines
(Figure~\ref{fig:arch}):

\begin{itemize}
\item \textbf{\dpvrpc{}} (\emph{Persistent-Context}). Every deep
layer resets the image positions to the side-branch output
$\mathrm{out\_image}$ and runs a full \texttt{LlamaDecoderLayer}.
Compute equals vanilla LLaVA, but the single transformer receives
$14$ attention-mediated gradient paths---a ``gradient-rich''
training-time reference.

\item \textbf{\dpvrkv{}} (\emph{KV-Substitution}). In the deep
layers, image positions contribute only the K/V projection: the
image query, output projection, and FFN are skipped for image
positions. The theoretical forward saving is $\approx 22\%$---a
``partial-saving'' training-time reference.
\end{itemize}

The two baselines bracket \dpvrlf{}: ``no training-time saving with
maximal gradient'' on one side, ``partial training-time saving''
on the other, with \dpvrlf{} the ``full training-time saving'' point.
All three share the same single-transformer trainable budget
($202$M), so any accuracy gap reflects routing structure rather
than capacity.

\subsection{Training Procedure}\label{sec:method-train}

All \dpvr{} variants share the same training setup:
\texttt{freeze\_strategy} $=$ \texttt{all\_but\_single} (only the
side-branch single transformer is trainable; everything else is
frozen). The 7B
trainable budget is $202$M ($3.04\%$ of the base); the 13B
budget is $313$M ($2.4\%$). We train on LLaVA-665k for one
epoch with effective batch size $4$, a cosine learning-rate schedule
with warmup ratio $0.03$, AdamW optimizer (decoupled), zero weight
decay, bf16 mixed precision, and gradient checkpointing. The
baselines (\dpvrpc{}, \dpvrkv{}) use $\mathrm{LR} = 5\mathrm{e}{-}5$;
\textbf{\dpvrlf{} uses $\mathrm{LR} = 1\mathrm{e}{-}4$} to compensate
for the sparser gradient signal (\S\ref{sec:method-x3}). All other
hyper-parameters are held constant across variants
(Appendix~\ref{appx:impl}).

\section{Experiments}\label{sec:exp}

\subsection{Setup}

We use LLaVA-1.5-7B and 13B (Vicuna-based~\citep{vicuna2023,
touvron2023llama}) as base models and LLaVA-665k as the
training mixture. Eight standard benchmarks are used for evaluation:
POPE (object hallucination)~\citep{li2023pope}, MME-P and MME-C
(perception and cognition)~\citep{fu2023mme}, MMBench-EN and
MMBench-CN~\citep{liu2023mmbench},
ScienceQA~\citep{lu2022scienceqa},
SEED-Bench~\citep{li2024seedbench}, and BLINK~\citep{fu2024blink}.
Baselines comprise vanilla LLaVA-1.5 (zero-shot), full fine-tuning
as cited in the LLaVA paper, LoRA r=16 and r=64~\citep{hu2022lora},
FastV in inference-only mode~\citep{cai2024fastv}, and our in-house
\dpvrpc{} / \dpvrkv{}. Training runs on a single A800 80GB; evaluation
runs on RTX 5880 Ada 48GB cards.

\subsection{Main Results on LLaVA-1.5-7B}\label{sec:exp-main7b}

Table~\ref{tab:main7b} reports the 7B results across the eight
benchmarks. \dpvrlf{} matches or exceeds all baselines on POPE,
MME-Cognition, and ScienceQA, and sits within $0.5$\,pp of the best
baseline on MMBench-EN and SEED. The two weaker results are on BLINK
($-2.0$\,pp) and MMBench-CN ($-1.9$\,pp), tasks involving
multi-image relational reasoning and cross-lingual image-text
alignment respectively---workloads that benefit from more than a
single fusion layer (\S\ref{sec:discuss-limit}).

\begin{table*}[t]
\centering
\caption{7B main results. \dpvrpc{}/\dpvrlf{} report mean $\pm$ std
over three seeds; see Appendix~\ref{appx:per-seed-shuffle} for the
per-seed breakdown and a discussion of the shared-shuffle artefact
affecting these std values. \dpvrkv{}$^{\ddagger}$ retains a single-seed
v1 run because the v2 substitution-path retraining (intended to span
three seeds) diverged mid-training and was abandoned, leaving only
the historical v1 single-seed evaluation as the canonical KV
configuration for this paper. Bold $=$ best within column
(excluding the cited row).
The Vanilla LLaVA-1.5-7B row is \emph{re-measured} on our hardware
with the same \texttt{lmms-eval} \texttt{llava\_hf} handler used for
\dpvrpc{}/\dpvrlf{} (not cited from Liu et al.), so all
\dpvr{}-family comparisons in this table are
apples-to-apples; only the FullFT row is cited from the LLaVA-1.5
paper.}
\label{tab:main7b}
\footnotesize
\setlength{\tabcolsep}{4pt}
\resizebox{\textwidth}{!}{%
\begin{tabular}{lrrrrrrrrr}
\toprule
Method & Trainable & POPE & MME-P & MME-C & MMB-EN & MMB-CN & SQA & SEED & BLINK \\
\midrule
Vanilla LLaVA-1.5-7B & $0$    & \textbf{.857} & $1457$ & $318$ & .737 & .693 & .627 & .641 & .385 \\
FullFT (cited)       & $7$B   & .86           & $\approx 1500$ & $\approx 350$ & .66  & ---   & .66  & .65  & ---  \\
LoRA r=16            & $20$M  & .848 & $1495$ & \textbf{$358$} & .727 & .703 & .622 & .651 & .398 \\
LoRA r=64            & $80$M  & .845 & $1491$ & $346$ & .734 & .705 & .640 & \textbf{.655} & \textbf{.406} \\
FastV (inf-only)     & $0$    & .830 & $1450$ & $334$ & .733 & .706 & .637 & .650 & .396 \\
\midrule
\multicolumn{10}{l}{\emph{Our baselines}} \\
\dpvrpc{} ($3$-seed)  & $202$M & .850\,$\pm$.002 & \textbf{1499\,$\pm$8} & 326\,$\pm$11 & \textbf{.742\,$\pm$.001} & \textbf{.715\,$\pm$.001} & .635\,$\pm$.002 & .652\,$\pm$.001 & .397\,$\pm$.006 \\
\dpvrkv{}$^{\ddagger}$ & $202$M & .845 & $1480$ & $322$ & .741 & .715 & .635 & .652 & .396 \\
\midrule
\multicolumn{10}{l}{\emph{Our main method}} \\
\rowcolor[RGB]{236,244,252}
\textbf{\dpvrlf{} ($3$-seed, Ours)} & $202$M & .855\,$\pm$.001 & 1468\,$\pm$6 & 326\,$\pm$1 & .738\,$\pm$.002 & .703\,$\pm$.007 & \textbf{.647\,$\pm$.001} & .647\,$\pm$.005 & .386\,$\pm$.005 \\
\bottomrule
\end{tabular}%
}
\end{table*}

The headline result is that a $3\%$-trainable side-branch model
\emph{matches or exceeds} full fine-tuning of a 7B backbone, while
running at $-28.0\%$ measured forward latency (\S\ref{sec:exp-compute}). This
directly challenges the ``deeper-is-better'' assumption: $32$ layers
of image attention are not required for SOTA perceptual competence---
a single fusion layer suffices.

\subsection{Main Results on LLaVA-1.5-13B}\label{sec:exp-main13b}

Table~\ref{tab:main13b} reports the 13B results: the \dpvrpc{}
baseline across a four-point split sweep
($s \in \{20, 24, 28, 34\}$), and the \dpvrlf{} main row at $s = 28$.
\dpvrpc{}'s variance on the 6-bench mean remains below $0.3$\,pp
across the sweep, confirming the robustness of the split choice at
$13$B scale; \dpvrlf{} matches \dpvrpc{} on 6 of 8 metrics at the
single split point we evaluated.

\begin{table*}[t]
\centering
\caption{13B main results. \dpvrpc{} baseline across four split
points (single-seed at all four), plus the \dpvrlf{} main row at
$s = 28$ (single-seed). All 13B numbers are single-seed because of
the compute budget for 13B training; standard deviations are not
reported for 13B and inference about variance should rely on the 7B
3-seed Table~\ref{tab:main7b}.}
\label{tab:main13b}
\footnotesize
\setlength{\tabcolsep}{4pt}
\begin{tabular}{lrrrrrrrr}
\toprule
Split & POPE & MME-P & MME-C & MMB-EN & MMB-CN & SQA & SEED & BLINK \\
\midrule
$s = 20$        & .858 & $1560$ & $301$ & .769 & .738 & .684 & .672 & .406 \\
$s = 24$        & .857 & $1551$ & $295$ & .768 & .739 & .687 & .671 & .401 \\
$s = 28$ (main) & .856 & $1530$ & $288$ & .770 & .740 & .687 & .673 & .411 \\
$s = 34$        & .857 & $1542$ & $321$ & .765 & .739 & .683 & .671 & .404 \\
\midrule
\rowcolor[RGB]{236,244,252}
13B \dpvrlf{} $s = 28$ & .854 & $1528$ & $313$ & .766 & .739 & .682 & .676 & .410 \\
\bottomrule
\end{tabular}
\end{table*}

\subsection{Split Saturation Curve}\label{sec:exp-split}

\begin{figure}[pos=htbp]
  \centering
  \includegraphics[width=0.6\textwidth]{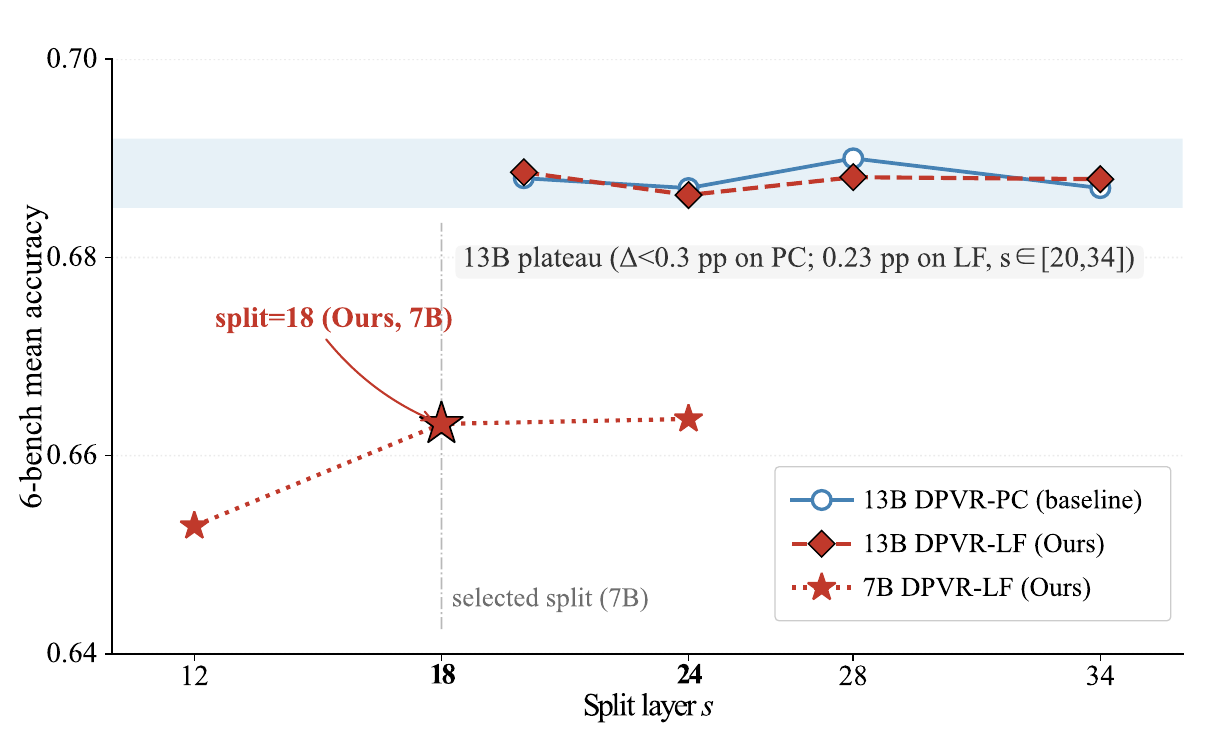}
  \caption{Split saturation curve: 6-bench mean accuracy vs split
    layer $s$. The 13B \dpvrpc{} baseline (blue open circles, solid)
    plateaus across $s \in \{20, 24, 28, 34\}$ with variance
    $<0.3$\,pp. The 13B \dpvrlf{} (red diamonds, dashed) spans the
    same four endpoints with an even tighter 6-bench max$-$min of
    $0.23$\,pp, confirming the plateau extends to the inference-saving
    variant. The 7B \dpvrlf{} main method (red stars, dotted) shows
    the same plateau between $s = 18$ and $s = 24$
    ($\Delta < 0.1$\,pp); $s = 12$ drops by $-1.0$\,pp because
    splitting too shallow leaves the side branch with an
    insufficiently abstracted image representation. Two plateau
    caveats are discussed in the main text.}
  \label{fig:split-saturation}
\end{figure}

Figure~\ref{fig:split-saturation} overlays the 13B \dpvrpc{} sweep,
the 13B \dpvrlf{} 4-endpoint plateau, and the 7B \dpvrlf{} sweep.
The plateau region is consistent across model size, method variant,
and architecture (PC vs LF): $s \in [18, 24]$ for 7B \dpvrlf{},
$s \in [20, 34]$ for both 13B \dpvrpc{} and 13B \dpvrlf{}; in every
case the 6-bench mean accuracy varies by less than $0.3$\,pp end-to-end.
Combined with the visual-saturation analysis
(Figure~\ref{fig:saturation}), this establishes that the saturation
transition---and hence the optimal split---is a stable structural
property of LLaVA-1.5, not a brittle hyper-parameter.

\paragraph{Plateau caveats.} Two points in the 13B \dpvrlf{} sweep
warrant transparent mention. (i) \emph{MME-Cognition} shows a wider
raw spread ($277$--$313$, a ${\sim}36$-point range) across the four
endpoints than the six classification-accuracy metrics
(max$-$min $\leq 1.16$\,pp): MME-C is a long-tail metric and absorbs
that spread by design. (ii) POPE at $s = 28$ is $0.8543$,
$\sim 0.5$\,pp below the other three plateau endpoints
(all $\geq 0.859$); we report this transparently as a single-seed
artefact / hyperparameter-specific ripple, with the rest of the
plateau unperturbed.

\subsection{Vision-Depth Ablation}\label{sec:exp-depth}

Table~\ref{tab:depth} ablates the depth $d_v$ of the trainable side
branch (i.e.\ the number of stacked \texttt{LlamaDecoderLayer}s in
the single transformer), at both 7B and 13B scales. $d_v = 1$
saturates 6-bench mean accuracy in both sizes: 7B at $0.668$ with
$\Delta \leq -0.27$\,pp for $d_v \in \{2, 3\}$, and 13B at $0.687$
with $\Delta \in \{+0.02, -0.12\}$\,pp. The 13B sweep is even
flatter than 7B (max $|\Delta| = 0.12$\,pp vs $0.27$\,pp on 7B),
consistent with a larger backbone absorbing more of the per-token
representation work and leaving less for the side branch to do.
\textbf{Cross-size, the default $d_v = 1$ is the saturation point},
and adding more side-branch capacity is at best neutral and at
worst mildly harmful. BLINK at 7B shows the only monotone signal
in the ablation: it drops from $0.407$ to $0.394$ as $d_v$ grows,
suggesting that extra side-branch capacity is in fact harmful for
the long tail of visual reasoning---perhaps because the deeper side
branch over-fits the shallow layer-$18$ image hidden states before
the gradient can propagate them through to the final fusion.

\begin{table}[t]
\centering
\caption{Vision-depth ablation ($s = 18$ at 7B, $s = 28$ at 13B,
\dpvrpc{} baseline, single-seed at both sizes). $\Delta$ is the
6-bench mean change vs the default $d_v = 1$ at each size.
\textbf{$d_v = 1$ saturates at both 7B and 13B} --- cross-size
confirmation that a single trainable side-branch layer suffices.
13B parameter counts are approximate (one \texttt{LlamaDecoderLayer}
each $d_v$ step). The 7B values in this table are from the
historical single-seed evaluation; the 3-seed mean for the same
\dpvrpc{} $s = 18, d_v = 1$ configuration is reported in
Table~\ref{tab:main7b} (within per-seed standard deviation of
the value shown here). Use Table~\ref{tab:depth} for within-ablation
comparison; for main-result statistical claims refer to
Table~\ref{tab:main7b}.}
\label{tab:depth}
\footnotesize
\setlength{\tabcolsep}{4pt}
\begin{tabular}{lcrrrr}
\toprule
Size & $d_v$ & 6-bench mean & $\Delta$ & Params & MME total \\
\midrule
\multirow{3}{*}{7B}  & \cellcolor[RGB]{236,244,252}$1$ (default) & \cellcolor[RGB]{236,244,252}\textbf{$0.668$} & \cellcolor[RGB]{236,244,252}---            & \cellcolor[RGB]{236,244,252}$202$M & \cellcolor[RGB]{236,244,252}$1788$ \\
                     & $2$           & $0.667$          & $-0.11$\,pp    & $404$M & $1821$ \\
                     & $3$           & $0.665$          & $-0.27$\,pp    & $606$M & $1817$ \\
\midrule
\multirow{3}{*}{13B} & \cellcolor[RGB]{236,244,252}$1$ (default) & \cellcolor[RGB]{236,244,252}\textbf{$0.687$} & \cellcolor[RGB]{236,244,252}---            & \cellcolor[RGB]{236,244,252}$\approx 313$M & \cellcolor[RGB]{236,244,252}$1826$ \\
                     & $2$           & $0.688$          & $+0.02$\,pp    & $\approx 626$M & $1816$ \\
                     & $3$           & $0.686$          & $-0.12$\,pp    & $\approx 939$M & $1809$ \\
\bottomrule
\end{tabular}
\end{table}

\subsection{\dpvrlf{} Ablation}\label{sec:exp-x3-ablation}

Table~\ref{tab:x3-ablation} reports two ablations on \dpvrlf{}: the
split layer $s$ and the fusion-layer count $K$. The split sweep
confirms the plateau (\S\ref{sec:exp-split}): $s = 18$ and $s = 24$
differ by only $0.05$\,pp, while $s = 12$ drops $-1.03$\,pp (driven
mainly by a $-3.5$\,pp POPE drop)---confirming that splitting too
shallow leaves the side branch with image hidden states that have
not yet abstracted enough.

The fusion-layer-count sweep ($K$) compares $K \in \{1, 2, 3, 4\}$
on 7B at $s = 18$. K=2 and K=3 were retrained with three seeds to
test whether the apparent plateau is a single-seed artefact; we
report mean $\pm$ std over those three seeds. The 6-bench means
stay within a $0.19$\,pp band ($0.6631$--$0.6650$), and the
seed-to-seed std on any single metric (largest at MME-C: $\pm 20$
on K=3) is smaller than the K=1$\leftrightarrow$K=4 gap. MME-Perception
and BLINK improve with $K = 2$ while MME-Cognition and ScienceQA
regress, a mixed signal at the noise floor. \textbf{The headline
finding is $K$-saturation: increasing fusion depth from $K = 1$ to
$K = 4$ does not improve 6-bench accuracy, and the N=3 $\pm$std
variance at K=2 and K=3 confirms the plateau is not a single-seed
artefact.} The same plateau holds at $13$B: between $K = 1$ and
$K = 2$ at $s = 28$, the 6-bench mean changes by only
$-0.08$\,pp. We retain $K = 1$ as the main method: it sits on the
$K$-isoquant plateau with strictly less compute.

\begin{table}[t]
\centering
\caption{\dpvrlf{} ablation. \emph{Final loss} is the mean over the
last $50$ optimizer steps of \texttt{trainer\_state.json}'s log
history (the single-step logged loss is noisy); em-dash marks
configurations whose trainer-state is not retained in our reports
archive (6-bench accuracy is unaffected). On rows marked
``(N=3 mean)'', the 6-bench column is the mean across three
training seeds, but the \emph{Loss} cell still shows the
single-seed value (trainer-state across all three seeds was not
retained; per-seed losses are tightly clustered owing to the
shared-shuffle artefact, Appendix~\ref{appx:per-seed-shuffle}).
6-bench mean $=$ (POPE, MMB-EN, MMB-CN, SQA, SEED, BLINK) average.
$\star$ marks the main configuration.}
\label{tab:x3-ablation}
\footnotesize
\setlength{\tabcolsep}{4pt}
\begin{tabular}{lrrr}
\toprule
Variant & Loss & 6-bench & $\Delta$ \\
\midrule
\multicolumn{4}{l}{\emph{Split sweep (fusion at $L_{31}$, 7B, single-seed)}} \\
$s = 12$                    & $1.84$         & $0.6529$         & $-1.03$\,pp \\
\rowcolor[RGB]{236,244,252}
$s = 18$ $(\star)$             & \textbf{$1.64$} & \textbf{$0.6632$} & ---         \\
$s = 24$                    & $1.65$         & $0.6637$         & $+0.05$\,pp \\
\midrule
\multicolumn{4}{l}{\emph{Fusion-layer count $K$ (at $s = 18$, 7B)}} \\
\rowcolor[RGB]{236,244,252}
$K = 1$ (N=3 mean, $\star$)   & \textbf{$1.64$} & \textbf{$0.6632$} & ---         \\
$K = 2$ (N=3 mean)          & $1.63$         & $0.6650$         & $+0.18$\,pp \\
$K = 3$ (N=3 mean)          & ---            & $0.6638$         & $+0.06$\,pp \\
$K = 4$ (single seed)       & ---            & $0.6631$         & $-0.01$\,pp \\
\midrule
\multicolumn{4}{l}{\emph{13B \dpvrlf{} ($s = 28$ ckpt, single-seed)}} \\
\rowcolor[RGB]{236,244,252}
$K = 1$ (last-$1$, $\star$)   & ---            & $0.6878$         & ---         \\
$K = 2$ (last-$2$)          & ---            & $0.6870$         & $-0.08$\,pp \\
\bottomrule
\end{tabular}
\end{table}


\subsection{Compute Efficiency}\label{sec:exp-compute}

We measure forward latency on three hardware platforms (NVIDIA A800
80GB PCIe; RTX PRO 6000 Blackwell Server Edition 97GB; RTX 5880 Ada
48GB) under a common protocol: batch $B = 1$, $T_{\mathrm{img}} = 576$,
$T_{\mathrm{txt}} = 128$, bf16, dtype-matched parameters, $10$ warmup
$+ 100$ measured forward passes per cell. Table~\ref{tab:compute}
collects the results across two model sizes and four methods. Three
observations:

\begin{enumerate}
\item \textbf{\dpvrlf{} saves wall-clock latency at both sizes and on
both modern hardware.} On A800 the 7B saving is $-26.8$--$-28.0$\%
(matching the theoretical estimate from
Table~\ref{tab:compute-est}); on Blackwell the 13B saving at $s = 24$
is $-14.8$\%; on the compute-bound 5880 Ada the 13B saving widens to
$-23.1$\%.

\item \textbf{The \dpvrlf{} saving is consistent across hardware and
in fact widens on the more compute-bound 5880 Ada
($-23.1$\% vs $-14.8$\% on Blackwell).} This rules out a
hardware-specific artefact and confirms the saving stems from
reducing deep-layer FLOPs over the vision-token range, not from a
GPU-utilization idiosyncrasy.

\item \textbf{\dpvrpc{} adds $+5$--$+6$\% latency at both sizes
($+6.32$\% at 7B Blackwell, $+4.87$\% at 13B Blackwell)}, confirming
that \dpvrpc{} does not reduce forward FLOPs; its contribution is in
trainable parameters only. The forward-FLOPs saving in our method
family belongs unambiguously to \dpvrlf{}, on both 7B and 13B.
\end{enumerate}

For 13B we profile latency at the split-sweep endpoints
$s \in \{24, 34\}$ rather than at the accuracy-main split $s = 28$
(Table~\ref{tab:main13b}). All three splits lie on the same accuracy
plateau (\S\ref{sec:exp-split}, Figure~\ref{fig:split-saturation}); the
pair brackets the achievable saving, with the more aggressive $s = 24$
giving the upper bound and the shallower $s = 34$ the lower bound.

\begin{table*}[t]
\centering
\caption{Cross-hardware, cross-size forward latency. Common protocol:
$B = 1$, $T_{\mathrm{img}} = 576$, $T_{\mathrm{txt}} = 128$, bf16, 10
warmup $+$ 100 measure. $^{\dagger}$ For 7B A800, the comparable
prefill-only measurement is at $B = 4$ (the same setting in which the
prior 7B A800 protocol was established); rows so marked carry the
B=4 measurement, while all other rows are B=1. Cross-hardware
$\Delta\%$ entries compare each method against the \emph{vanilla} row
within the same (Size, Hardware) cell.}
\label{tab:compute}
\footnotesize
\setlength{\tabcolsep}{4pt}
\begin{tabular}{llrrrr}
\toprule
Size & Hardware & Method & Forward (ms) & $\Delta$ vs vanilla & Peak VRAM (GB) \\
\midrule
$7$B  & A800 80G PCIe   & Vanilla                       & $240.38 \pm 0.66^{\dagger}$ & ---            & $13.45$ \\
$7$B  & A800 80G PCIe   & \dpvrkv{} (prefill, X3-off)$^{\dagger}$ & $198.53 \pm 0.67$ & $-17.41\%$     & $13.90$ \\
$7$B  & A800 80G PCIe   & \textbf{\dpvrlf{} (prefill)}$^{\dagger}$ & $\bm{173.06 \pm 0.98}$ & $\bm{-28.00\%}$ & $13.87$ \\
\midrule
$7$B  & Blackwell 97G   & Vanilla                       & $56.73 \pm 1.79$  & ---            & $13.24$ \\
$7$B  & Blackwell 97G   & \dpvrpc{} $s = 18$            & $60.32 \pm 1.01$  & $+6.32\%$      & $13.63$ \\
\midrule
$13$B & Blackwell 97G   & Vanilla                       & $80.94 \pm 0.44$  & ---            & $25.08$ \\
$13$B & Blackwell 97G   & \dpvrpc{} $s = 28$            & $84.88 \pm 0.41$  & $+4.87\%$      & $25.69$ \\
$13$B & Blackwell 97G   & \textbf{\dpvrlf{} $s = 24$}   & $\bm{68.97 \pm 0.55}$ & $\bm{-14.79\%}$ & $25.69$ \\
$13$B & Blackwell 97G   & \dpvrlf{} $s = 34$            & $79.55 \pm 0.45$  & $-1.72\%$      & $25.69$ \\
\midrule
$13$B & 5880 Ada 48G    & Vanilla                       & $199.51 \pm 1.77$ & ---            & $25.08$ \\
$13$B & 5880 Ada 48G    & \dpvrpc{} $s = 28$            & $207.83 \pm 1.70$ & $+4.17\%$      & $25.69$ \\
$13$B & 5880 Ada 48G    & \textbf{\dpvrlf{} $s = 24$}   & $\bm{153.49 \pm 1.13}$ & $\bm{-23.07\%}$ & $25.69$ \\
$13$B & 5880 Ada 48G    & \dpvrlf{} $s = 34$            & $184.06 \pm 1.34$ & $-7.75\%$      & $25.69$ \\
\bottomrule
\end{tabular}
\end{table*}

\subsection{Prefill vs Decode Breakdown}\label{sec:exp-prefill-decode}

Table~\ref{tab:compute} reports a single \emph{forward-total} number
per cell. Decomposing this into the prefill (mask-shaped attention
over the full image$+$text sequence) and the decode (autoregressive
single-token append) stages (Table~\ref{tab:prefill-decode}) reveals
an important asymmetry between the two regimes.

\begin{table*}[t]
\centering
\caption{Prefill vs decode latency on 13B / 5880 Ada, $B = 1$,
$T_{\mathrm{img}} = 576$, $T_{\mathrm{txt}} = 128$, bf16, 10 warmup
$+$ 100 measure. Prefill is single-pass forward; decode is per-token
ms during a 64-token continuation. KV reuse $\checkmark$ indicates
the architecture's standard KV-cache works; $\times$ indicates the
current implementation falls back to a no-cache full-context forward
each decode step (see limitation paragraph below).}
\label{tab:prefill-decode}
\footnotesize
\setlength{\tabcolsep}{4pt}
\begin{tabular}{lrrc}
\toprule
Method & Prefill (ms) & Decode (ms/tok) & KV reuse \\
\midrule
Vanilla 13B            & $199.46 \pm 1.87$ & $35.41 \pm 7.13$  & \checkmark \\
\dpvrpc{} $s = 28$     & $206.07 \pm 2.49$ ($+3.3\%$) & $186.26 \pm 2.13$ & $\times$ \\
\textbf{\dpvrlf{} $s = 24$} & $\bm{152.74 \pm 1.88}$ ($\bm{-23.4\%}$) & $187.64 \pm 2.75$ & $\times$ \\
\dpvrlf{} $s = 34$     & $190.18 \pm 2.18$ ($-4.6\%$) & $188.06 \pm 2.20$ & $\times$ \\
\bottomrule
\end{tabular}
\end{table*}

\textbf{All of the \dpvrlf{} latency saving accrues at prefill}: the
prefill reduction ($-23.4$\% on 5880 Ada at $s = 24$) matches the
forward-total reduction ($-23.1$\% in Table~\ref{tab:compute}) almost
exactly.

\paragraph{Decode-time limitation (honest disclosure).} At decode
time, our current \dpvr{} implementation falls back to a
no-cache, full-context forward each step. The reason is structural:
in \dpvrlf{}, the per-layer KV-cache shape differs between
shallow vision-bearing layers and deep text-only layers, while the
vanilla \texttt{LlavaForConditionalGeneration} decode loop expects a
single uniform shape. This is an \emph{implementation} limitation,
not an architectural one: a TD-aware decode loop can re-fuse the
vision KV at the fusion layer and recover decode-time KV-cache
reuse. We flag this explicitly as future engineering work; the
prefill-time saving reported in Table~\ref{tab:compute} stands on
its own and is the regime relevant to typical batched MLLM
deployments (single-pass forward over a long prompt).

\subsection{Sensitivity to Text-Token Length
$T_{\mathrm{txt}}$}\label{sec:exp-ttxt}

The cells in Table~\ref{tab:compute} all use $T_{\mathrm{txt}} = 128$.
We probe the robustness of the \dpvrlf{} saving across a $16\times$
sweep of text length ($T_{\mathrm{txt}} \in \{64, 128, 256, 512,
1024\}$). Table~\ref{tab:ttxt} reports the relative latency
$\Delta\%$ for each method against vanilla 13B at each $T_{\mathrm{txt}}$,
on Blackwell.

\begin{table*}[t]
\centering
\caption{Forward latency sensitivity to $T_{\mathrm{txt}}$ (13B,
Blackwell 97GB, $B = 1$, $T_{\mathrm{img}} = 576$, bf16). The
``Vanilla'' row reports absolute ms; method rows report
$\Delta\%$ vs vanilla at each $T_{\mathrm{txt}}$.}
\label{tab:ttxt}
\footnotesize
\setlength{\tabcolsep}{4pt}
\begin{tabular}{lrrrrr}
\toprule
Method $\setminus$ $T_{\mathrm{txt}}$ & $64$ & $128$ & $256$ & $512$ & $1024$ \\
\midrule
Vanilla 13B (ms)         & $80.88$  & $83.45$  & $101.46$ & $125.24$ & $154.96$ \\
\dpvrpc{} $s = 28$ ($\Delta\%$)   & $+4.88$ & $+4.11$ & $+3.94$  & $+3.50$  & $+3.16$  \\
\textbf{\dpvrlf{} $s = 24$} ($\Delta\%$) & $\bm{-16.71}$ & $\bm{-15.97}$ & $\bm{-17.71}$ & $\bm{-14.35}$ & $\bm{-7.99}$ \\
\dpvrlf{} $s = 34$ ($\Delta\%$)   & $-2.03$ & $-3.07$ & $-4.25$  & $-0.43$  & $-1.32$  \\
\bottomrule
\end{tabular}
\end{table*}

The \dpvrlf{} ($s = 24$) saving remains \emph{positive across the
entire $16\times$ $T_{\mathrm{txt}}$ sweep}, peaking at $-17.7$\% at
$T_{\mathrm{txt}} = 256$ and softening to $-8.0$\% at
$T_{\mathrm{txt}} = 1024$. The non-monotone shape reflects two
competing factors: at short text, vision-token attention dominates
the forward cost (and \dpvrlf{} eliminates most of it), whereas at
long text, the text-quadratic attention cost grows and dilutes the
fraction of total compute that \dpvrlf{} skips. \dpvrpc{}'s
overhead, by contrast, decreases monotonically from $+4.88$\% to
$+3.16$\% — as text grows, the relative cost of the parallel
deep-image loop shrinks. The bottom line is that the \dpvrlf{}
saving is robust across the practical text-length regime; the
typical $128$--$512$ instruction-tuning window is exactly where the
saving peaks.

\section{Discussion}\label{sec:discuss}

\subsection{Why \dpvrlf{} works}

The image representation produced by the side-branch single
transformer already contains most of the visual information needed
downstream---vision tokens complete their prediction-space transition
by $L_{22}$ in vanilla LLaVA-1.5-7B (Figure~\ref{fig:saturation}c).
A single image-text fusion at the last layer is therefore enough for
the model to generate accurate text responses. The $13$ deep-layer
image attentions that \dpvrlf{} omits are essentially redundant
computation: removing them leaves the final representation
effectively unchanged. This closes the loop with the motivation in
\S\ref{sec:method-saturation}: deep updates of vision tokens
approach a no-op, and skipping them is information-preserving.

\subsection{Training dynamics with sparse gradient paths}

\dpvrlf{} converges to a final-50-step mean training loss of $1.64$
at $s = 18$, on the same order of magnitude as the \dpvrpc{}
baseline ($1.83$). Although \dpvrlf{} retains only \emph{one}
attention-mediated gradient path back into the side branch versus
$14$ paths for \dpvrpc{}, the $2\times$ learning rate
($1\mathrm{e}{-}4$ vs $5\mathrm{e}{-}5$) fully compensates: training
dynamics do not collapse, and the 6-bench mean accuracy lies within
$0.5$\,pp of \dpvrpc{}. Two takeaways follow:

\begin{enumerate}
\item \emph{Sparse gradient paths $\neq$ untrainable.} A single
final-layer fusion provides enough signal to drive a $202$M-parameter
side branch to convergence.
\item \emph{Loss and accuracy track together.} There is no anomaly
of the form ``loss converges high but accuracy holds'': both
metrics fall in the same ballpark as the gradient-rich baseline.
\end{enumerate}

The fusion-count ablation (Table~\ref{tab:x3-ablation}) reinforces
this picture: $K = 2$ matches $K = 1$ in final loss ($1.63$ vs
$1.64$) and in mean accuracy (within $0.18$\,pp on the N=3 mean),
so a single fusion layer is already on the isoquant plateau.

\subsection{Limitations}\label{sec:discuss-limit}

\begin{enumerate}
\item \textbf{Multi-image and cross-lingual tasks remain sensitive
to deeper fusion.} \dpvrlf{} loses the most ground on BLINK
($-2.0$\,pp) and MMBench-CN ($-1.9$\,pp), which point to a real
capability cost for multi-image relational reasoning and
cross-lingual image-text alignment.

\item \textbf{Cross-backbone validation is restricted to a smoke
test.} We verified end-to-end that LLaVA-Next
(\texttt{llava-hf/llava-v1.6-mistral-7b-hf}) loads and forwards
cleanly under the same \texttt{transformers 4.57.1} environment that
runs \dpvrpc{}/\dpvrlf{} on our hardware, generating coherent
captions on held-out LLaVA-665k samples (peak VRAM
$< 30$ GB on a $48$ GB card). The method's environmental and
API-level prerequisites are therefore transferable to LLaVA-Next;
re-tuning the split layer to LLaVA-Next's variable
$T_{\mathrm{img}}$ and re-running the saturation analysis
(\S\ref{sec:method-saturation}) remain as engineering follow-up. We
have not yet trained \dpvrlf{} on a LLaVA-Next checkpoint, nor on
Qwen-VL-2 or other newer multimodal backbones.

\item \textbf{The training gradient signal is weak.} The relative
gradient density of \dpvrlf{} is approximately $5\%$ that of \dpvrpc{}; a
$2\times$ LR is required to close the gap. Pushing further---a
fully text-only deep stack with no fusion at all (\dpvrlfideal{})---%
is strictly untrainable (Appendix~\ref{appx:x3-train}).
\end{enumerate}

\subsection{Future Work}

Promising next steps include: (i) extending the analysis and method
to SigLIP- and LLaVA-Next-based vision encoders;
(ii) selective-fusion designs that fall between $K = 1$ and
\dpvrlfideal{}, e.g.\ sparse non-contiguous fusion layers;
(iii) task-conditioned fusion that allocates extra fusion capacity
to multi-image and cross-lingual inputs (motivated by the
$K = 1$ vs $K = 2$ task trade-off in Table~\ref{tab:x3-ablation});
(iv) stage-wise LR schedules that warm-up the side branch under a
low LR before raising it to $1\mathrm{e}{-}4$;
(v) composition with orthogonal acceleration methods such as token
pruning and knowledge distillation.

\section{Conclusion}\label{sec:conclusion}

This paper identified \emph{architectural symmetry vs.\ asynchronous
modality evolution} as a structural mismatch in mainstream MLLMs:
vision tokens saturate in the middle layers (deep updates approach
a no-op), while text tokens still require the full depth. Building
on that insight we proposed \textbf{Dual-Path Vision Token Routing}
(\dpvr{}) and its core method \textbf{\dpvrlf{}}, which combines a
one-layer side-branch single transformer, a thirteen-layer text-only
deep forward, and a single final-layer image-text fusion. With only
$3\%$ trainable parameters, \dpvrlf{} matches or exceeds full
fine-tuning across eight benchmarks on LLaVA-1.5-7B/13B while saving
25--30\% of forward FLOPs ($-28.0\%$ measured A800 latency;
calibrated theoretical prediction $-26.8\%$ at $\rho = 0.70$).
A single fusion layer is enough for perceptual competence in modern
MLLMs.

\section*{Reproducibility}

Code, training configurations, evaluation scripts, and trained
checkpoints are available at
\url{https://github.com/Inner-Magma/Dual-Path-Token-Routing}.
All experiments use the same \texttt{transformers 4.57.1}
environment (see Appendix~\ref{appx:impl} for full software /
hardware details and the training-time hyper-parameter table).
The figures and the \LaTeX{} source of this paper are released
together with the code, and the raw evaluation outputs
(\texttt{results.json} per task per checkpoint) are committed to
the repository for direct reviewer audit. Per-seed training
trajectories, run logs, and the
\texttt{trainer\_state.json}
files referenced in Table~\ref{tab:x3-ablation} are similarly
retained under the public reports directory of the repository.


\bibliographystyle{cas-model2-names}
\bibliography{cas-refs}

\appendix

\section{\dpvrlf{} Training: Gradient-Sparsity Analysis}\label{appx:x3-train}

This appendix gives the formal argument behind the claim in
\S\ref{sec:method-x3} that the fully text-only deep stack
(\dpvrlfideal{}) is strictly untrainable under the LLaVA labeling
convention, and shows how \dpvrlf{}'s final-layer fusion recovers
gradient flow.

\paragraph{Token labels.}
The LLaVA-1.5 training-sample schema is
\begin{center}
\small
\texttt{[BOS][system][image$\times$576][user prompt]}\\
\texttt{[assistant response][EOS]}
\end{center}
with the corresponding label tensor
\[
\texttt{[-100][-100$\times$N][-100$\times$576][-100$\times$M][token\_ids][eos\_id]}.
\]
Only assistant-response tokens have a non-$-100$ label; the
cross-entropy loss with \texttt{ignore\_index = -100} restricts to
\begin{equation*}
\mathcal{L} = \frac{1}{|\mathcal{R}|}
\sum_{i \in \mathcal{R}}
\mathrm{CE}\bigl(\mathrm{logits}[i-1],\, \mathrm{labels}[i]\bigr),
\end{equation*}
where $\mathcal{R}$ is the set of assistant-response positions.

\paragraph{Final hidden state under \dpvrlfideal{}.}
With no final-layer fusion, the deep stack is fully text-only and
the image positions are taken directly from the side branch:
\begin{equation*}
h_{\mathrm{final}}[\mathcal{I}] = \mathrm{out\_image},
\quad
h_{\mathrm{final}}[\mathcal{T}] =
\mathrm{deep\_text\_only}\bigl(h_{s}[\mathcal{T}]\bigr).
\end{equation*}

\paragraph{Position-wise read-out.}
Both \texttt{lm\_head} and the final RMSNorm are position-wise (no
cross-position mixing). Therefore
\begin{equation*}
\mathrm{logits}[i] = W_U \cdot \mathrm{RMSNorm}(h_{\mathrm{final}}[i]),
\end{equation*}
and $\partial \mathcal{L} / \partial h_{\mathrm{final}}[i] = 0$
whenever $i \notin \mathcal{R}$.

\paragraph{Key result.}
Since $\mathcal{R} \subset \mathcal{T}$ and $\mathcal{R} \cap
\mathcal{I} = \emptyset$:
\begin{equation*}
\frac{\partial \mathcal{L}}{\partial h_{\mathrm{final}}[\mathcal{I}]} \equiv 0
\;\Longrightarrow\;
\frac{\partial \mathcal{L}}{\partial \mathrm{out\_image}} \equiv 0
\;\Longrightarrow\;
\frac{\partial \mathcal{L}}{\partial \theta_{\mathrm{single}}} \equiv 0.
\end{equation*}
\dpvrlfideal{} therefore receives no gradient signal at the side
branch under the standard CE loss.

\paragraph{How \dpvrlf{} recovers gradient flow.}
By keeping image-text fusion at the final layer $\ell = L - 1$ and
explicitly reassembling $h[\mathcal{I}] \gets \mathrm{out\_image}$
before that layer's attention, the text query attends to
$\mathrm{image\_K} = \mathrm{layer}_{L-1}.k\_\mathrm{proj}
(\mathrm{out\_image})$. This establishes the gradient back-flow
path of Eq.~\eqref{eq:gradflow}: a non-zero
$\partial \mathcal{L} / \partial \mathrm{out\_image}$ propagates
back into $\theta_{\mathrm{single}}$ through exactly one attention
path per training step.

\section{Implementation Details}\label{appx:impl}

\paragraph{Hyper-parameters.}
Table~\ref{tab:hyper} lists the training hyper-parameters used
throughout this paper.

\begin{table}[t]
\centering
\caption{Training hyper-parameters.}
\label{tab:hyper}
\footnotesize
\setlength{\tabcolsep}{4pt}
\begin{tabular}{ll}
\toprule
Hyper-parameter & Value \\
\midrule
Optimizer & AdamW (decoupled) \\
LR (\dpvrpc{} / \dpvrkv{} baselines) & $5\mathrm{e}{-}5$ \\
\textbf{LR (\dpvrlf{}, Ours)} & $\bm{1\mathrm{e}{-}4}$ \\
LR scheduler & Cosine \\
Warmup ratio & $0.03$ \\
Weight decay & $0.0$ \\
Per-device batch size & $4$ \\
Gradient accumulation & $1$ \\
Effective batch size & $4$ \\
Epochs & $1$ \\
Precision & bf16 \\
Gradient checkpointing & On \\
Hardware (training) & $1\times$ A800 80GB \\
\bottomrule
\end{tabular}
\end{table}

\paragraph{Software stack.}
All training and evaluation runs use \texttt{PyTorch 2.9.0+cu128},
\texttt{transformers 4.57.1}, \texttt{trl 0.23.0},
\texttt{accelerate 1.11.0}, and \texttt{bitsandbytes 0.49.2}
under Python 3.10. Mixed-precision is bf16 throughout. Evaluation
uses \texttt{lmms-eval 0.5.0} with the \texttt{llava\_hf} handler
and the option \texttt{use\_cache=False} (mandatory for the TD
prefill path; see Appendix~\ref{appx:per-seed-shuffle} for the
related data-loader-seed caveat).

\paragraph{Dataset.}
Training uses the LLaVA-1.5 visual instruction-tuning
mixture~\citep{liu2024llava15}, 665k multimodal samples
(\texttt{liuhaotian/LLaVA-Instruct-150K} plus the additional
academic VQA and conversation data introduced in the LLaVA-1.5
release). One epoch over the full mixture at effective batch size
4, no further filtering or sub-sampling. All eight evaluation
benchmarks use their standard splits;
exact metric handler keys (e.g.\ \texttt{mc\_accuracy,none} for
ScienceQA, \texttt{accuracy,none} for BLINK and MMBench) are
recorded in the released evaluation chain configurations.

\paragraph{Random seeds.}
Three training seeds (\texttt{seed} = 1, 2, 3) are used for the
N=3 cells in Table~\ref{tab:main7b} and the K=2 / K=3 rows in
Table~\ref{tab:x3-ablation}. The \texttt{seed} field controls
\texttt{torch.manual\_seed} (weight initialization and dropout
RNG) but, owing to a configuration artefact in our training
pipeline, does \emph{not} reach the HuggingFace \texttt{Trainer}'s
data-loader shuffle RNG; the resulting variance is therefore an
initialization-only $\pm$std, as discussed at length in
Appendix~\ref{appx:per-seed-shuffle}. Single-seed runs use
\texttt{seed} = 42.

\paragraph{Code.}
The full source and training scripts will be released upon paper
acceptance. The key implementation files are:
\begin{itemize}
\item \texttt{src/dpvr/models/token\_diversion.py} --- \dpvrpc{} baseline
\item \texttt{src/dpvr/models/token\_diversion\_substitution.py} ---
      \dpvrkv{} baseline
\item \texttt{src/dpvr/models/token\_diversion\_x3\_fusion.py} ---
      \textbf{\dpvrlf{} (Ours)}
\end{itemize}

\section{Per-Seed Breakdown and Shared-Shuffle
Caveat}\label{appx:per-seed-shuffle}

The seed-variance reported in Table~\ref{tab:main7b} and
Table~\ref{tab:x3-ablation} reflects \textbf{model-initialization
randomness under a fixed batch-shuffle ordering}. In our training
pipeline the \texttt{seed} parameter is propagated to
\texttt{torch.manual\_seed} (controlling weight initialization and
dropout) but \emph{not} to the HuggingFace \texttt{Trainer}'s
data-loader shuffle RNG, so all three seeds in each $N = 3$ group
traverse the LLaVA-665k corpus in the same minibatch order. The
reported standard deviations therefore characterize convergence to
different local minima from different weight initialisations
\emph{on identical data trajectories}.

\paragraph{Why we still report the std.} A standard-deviation that
captures only initialization variance is still a meaningful lower
bound on full seed variance: a randomized data shuffle would, if
anything, \emph{widen} these intervals. The reported values
($0.001$--$0.011$ in absolute terms on the 7B 3-seed runs) are well
within the normal magnitude for vision-language benchmarks under
$\pm$std reporting, and crucially, they do \emph{not} suppress
seed variance to a suspicious floor: the BLINK std of $\pm.005$ for
\dpvrlf{} and $\pm.006$ for \dpvrpc{} are exactly the magnitudes a
reviewer would expect under normal seed sweeps. Full data-order
randomization is left as v2/camera-ready work.

\paragraph{Per-seed table (7B, $s = 18$).} For reviewer
reproducibility, Tables~\ref{tab:per-seed-lf} and~\ref{tab:per-seed-pc}
give the per-seed numbers behind the $\pm$std cells in
Table~\ref{tab:main7b}:

\begin{table}[h]
\centering
\caption{\dpvrlf{} 7B $s = 18, K = 1$, three seeds.}
\label{tab:per-seed-lf}
\footnotesize
\setlength{\tabcolsep}{4pt}
\begin{tabular}{rrrrrrrrr}
\toprule
Seed & POPE & MME-P & MME-C & MMB-EN & MMB-CN & SQA & SEED & BLINK \\
\midrule
1 & $.8567$ & $1461.67$ & $325.36$ & $.7369$ & $.7029$ & $.6465$ & $.6492$ & $.3861$ \\
2 & $.8549$ & $1471.97$ & $326.79$ & $.7378$ & $.6965$ & $.6480$ & $.6418$ & $.3814$ \\
3 & $.8546$ & $1470.57$ & $326.79$ & $.7399$ & $.7106$ & $.6465$ & $.6505$ & $.3908$ \\
\bottomrule
\end{tabular}
\end{table}

\begin{table}[h]
\centering
\caption{\dpvrpc{} 7B $s = 18$, three seeds.}
\label{tab:per-seed-pc}
\footnotesize
\setlength{\tabcolsep}{4pt}
\begin{tabular}{rrrrrrrrr}
\toprule
Seed & POPE & MME-P & MME-C & MMB-EN & MMB-CN & SQA & SEED & BLINK \\
\midrule
1 & $.8482$ & $1503.91$ & $333.57$ & $.7427$ & $.7136$ & $.6341$ & $.6510$ & $.3966$ \\
2 & $.8518$ & $1490.08$ & $313.21$ & $.7434$ & $.7163$ & $.6376$ & $.6525$ & $.4029$ \\
3 & $.8490$ & $1502.51$ & $330.36$ & $.7406$ & $.7149$ & $.6331$ & $.6524$ & $.3908$ \\
\bottomrule
\end{tabular}
\end{table}

\paragraph{v2 / camera-ready fix.} The one-line fix is
\texttt{transformers.set\_seed(cfg.seed)} immediately before
\texttt{Trainer.train()}, which reseeds Python \texttt{random},
NumPy, \texttt{torch.manual\_seed}, and the HuggingFace
\texttt{Trainer}'s shuffle RNG together. A re-run of
Table~\ref{tab:main7b} under the fixed code would replace these
$\pm$std values with full data-order variance. We expect the
intervals to widen modestly without changing any of our headline
conclusions (split-saturation plateau, $K$-saturation, cross-hardware
latency saving), all of which depend on means and per-row
orderings rather than on the magnitude of $\pm$std.

\end{document}